\documentclass[runningheads]{llncs}
\usepackage[T1]{fontenc}
\usepackage{booktabs}
\usepackage{float}
\usepackage{graphicx,verbatim}
\usepackage[hidelinks, colorlinks,citecolor=blue]{hyperref}
\usepackage{tabularx}
\usepackage{booktabs}
\usepackage{amsmath}

\def\Plus{\texttt{++}}
\begin{document}
\title{REVEAL\Plus: Differentiable Phenotypic Grouping for Vision–Language Retinal Modeling of Alzheimer’s Disease Risk}
\titlerunning{REVEAL\Plus}
%

\author{Ethan Meidinger \inst{1} \and Seowung Leem \inst{2} \and Zeyun Zhao \inst{2} \and Ruogu Fang \inst{2}\thanks{Corresponding author: ruogu.fang@ufl.edu}}
%
\authorrunning{Meidinger et al.}
%
\institute{
University of Virginia, Charlottesville, VA, USA \and
J. Crayton Pruitt Family Department of Biomedical Engineering, Herbert Wertheim College of Engineering, University of Florida, Gainesville, FL, USA
}

\maketitle              
%
\begin{abstract}
The retina offers a noninvasive window into neurodegenerative disease, capturing subtle structural patterns associated with a risk of future cognitive decline. Vision–language alignment frameworks such as REVEAL have shown that pairing retinal fundus images with structured clinical risk narratives improves early prediction of Alzheimer’s disease (AD). A key design choice in these approaches is the use of phenotypic grouping, where individuals with similar risk profiles are treated as multi-positive pairs during contrastive learning. However, existing methods operationalize phenotypic similarity as a discrete construct, relying on hard group assignments that impose rigid supervision and decouple group formation from representation learning. We propose a continuous formulation of phenotypic structure within contrastive learning. Rather than assigning samples to fixed clusters, we model inter-subject similarity as a differentiable weighting function derived from intra-modality embedding similarities in both retinal images and risk profiles. These weights define soft multi-positive relationships through a continuous aggregation operator, enabling graded supervision that reflects the spectrum nature of disease risk. We further introduce a soft-target contrastive objective that jointly learns cross-modal alignment and phenotypic structure in an end-to-end manner. Evaluated on UK Biobank retinal imaging data for incident AD prediction, the proposed framework consistently outperforms discrete group-based contrastive learning and standard vision–language baselines. By treating phenotypic similarity as a learnable, continuous signal rather than a fixed grouping rule, our approach provides a principled and robust foundation for population-scale neurodegenerative risk modeling from multi-modal retinal and clinical data.

\end{abstract}

%
%
%
\section{Introduction}

\begin{figure}
  \centering
  \includegraphics[width=\linewidth]{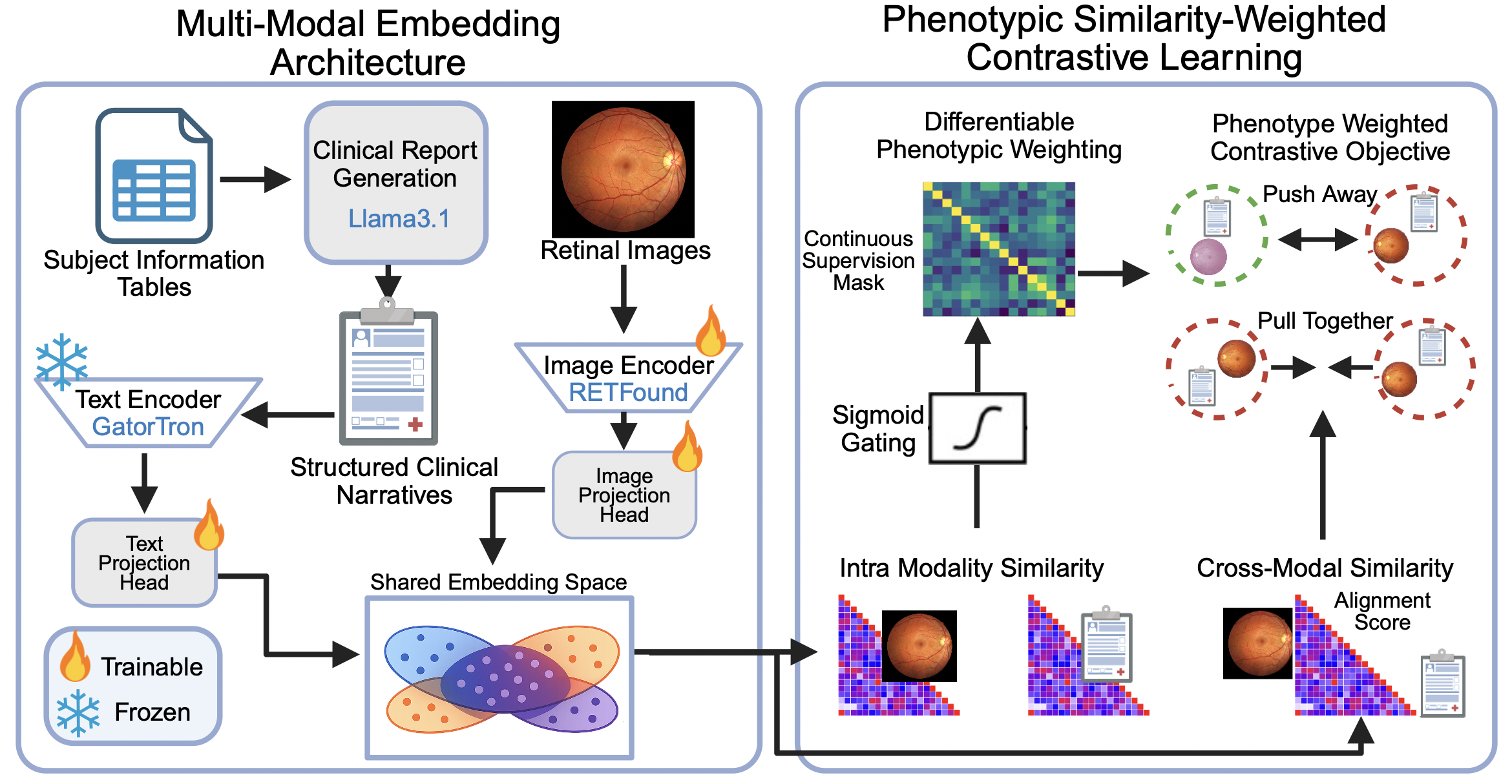}
  \caption{Architecture of the proposed differentiable phenotypic weighting framework for group-aware contrastive learning. Image and text embeddings are aligned via a similarity-weighted multi-positive contrastive loss, where continuous phenotypic weights replace hard grouping to model the heterogeneous spectrum of Alzheimer’s disease risk}
  \label{f}
\end{figure}

Alzheimer’s disease (AD) is a progressive neurodegenerative disorder characterized by a long preclinical phase during which pathological changes accumulate before clinical symptoms \cite{Chow2026NeurodevelopmentalOrigins}. Advances in brain imaging and plasma-based biomarkers have substantially improved the ability to detect disease-related pathology. However, these approaches may remain costly, invasive, or impractical for large-scale population screening. Complementary modalities that are scalable and noninvasive therefore, play an important role in early risk stratification. The retina has emerged as one such modality, as its structure and microvasculature share developmental and physiological links with the central nervous system and are associated with neurodegenerative and vascular processes relevant to AD \cite{Banna2024ImagingEyeBrain}. In parallel, systemic and lifestyle-related risk factors, including cardiometabolic health and sleep patterns, capture longitudinal exposures that contribute to dementia risk decades before diagnosis \cite{BuenoLopez2025CardiometabolicDementia,Suzgun2025SleepADRisk}. Rather than serving as standalone diagnostic markers, these signals offer complementary population-level information that may help characterize early disease susceptibility.

Recent advances in vision–language models (VLMs), has enabled joint representation learning across heterogeneous data modalities through contrastive alignment. Inspired by CLIP-style architectures, medical VLMs have increasingly been adapted to retinal imaging, leveraging large-scale pretraining to learn clinically meaningful visual representations \cite{Du2024RETCLIP,Wu2024MMRetinal,Zhou2023RETFound}. Building on this paradigm, the REVEAL framework aligned retinal fundus images with individualized clinical risk narratives derived from structured health data, enabling multi-modal modeling of neurodegenerative risk \cite{Leem2026REVEAL}. A central innovation of REVEAL was group-aware contrastive learning (GACL), which encouraged subjects with similar phenotypic profiles to act as multi-positive pairs during training. This strategy improved robustness to individual-level noise and promoted learning of shared disease-relevant structure, leading to improved downstream AD risk prediction compared with uni-modal and standard pairwise contrastive approaches.

Despite these advantages, phenotypic grouping in existing GACL formulations is constructed through discrete similarity thresholds, implicitly assuming that individuals belong to well-separated risk categories. From a biological perspective, however, neurodegenerative risk evolves along continuous and overlapping trajectories shaped by heterogeneous genetic, vascular, metabolic, and lifestyle factors. Individuals often exhibit partial similarity across multiple phenotypic axes rather than membership in a single homogeneous group. Hard group assignments may therefore introduce artificial boundaries that fail to reflect the graded and spectrum-like nature of disease vulnerability, while preventing the grouping process itself from adapting during representation learning.

In this work, we introduce a differentiable phenotypic weighting framework that treats inter-subject similarity as a continuous supervisory signal within multi-modal contrastive learning. Rather than relying on threshold-based clustering, similarity structures are computed directly from retinal image embeddings and clinical risk-profile embeddings and combined through a soft aggregation operator to produce continuous group-membership weights. These weights define a soft multi-positive contrastive objective in which supervision strength varies smoothly according to phenotypic proximity. By modeling phenotypic relationships as a differentiable attention-like process, the proposed framework enables joint learning of representation alignment and population-level structure, more faithfully capturing the continuous and heterogeneous biological variability underlying neurodegenerative risk.

Our contributions are three-fold:
\begin{itemize}

\item \textbf{Differentiable phenotypic weighting:}
We replace hard threshold-based grouping in group-aware contrastive learning with continuous phenotypic similarity weights derived from retinal and clinical embeddings, enabling smooth data-driven cohort modeling that better captures heterogeneous Alzheimer’s disease risk.

\item \textbf{Soft multi-positive contrastive learning:} 
We introduce a soft-target contrastive objective that incorporates phenotypic similarity into cross-modal alignment, enabling graded multi-positive supervision instead of binary pair assignments.

\item \textbf{State-of-the-art Alzheimer’s risk prediction from retinal imaging:}
We achieve new state-of-the-art performance on UK Biobank retinal imaging for incident Alzheimer’s disease prediction, outperforming existing vision–language and group-aware contrastive learning methods.

\end{itemize}

\section{Methods}

\subsection{Overview of REVEAL++}

REVEAL learns joint image–text representations of retinal fundus images and structured clinical reports under a group-aware contrastive objective. 
Given a minibatch of $N$ subjects, each subject $p$ is associated with a retinal image and a clinical report. 
Image and text encoders produce modality-specific embeddings, which are projected into a shared latent space. 
To incorporate phenotypic structure into contrastive supervision, we compute intra-modality similarity matrices that capture retinal image embedding and risk-profile similarity between subjects. 
These similarities are transformed into a differentiable phenotypic weighting mask $W \in [0,1]^{N\times N}$, which acts as a soft pairwise target matrix in a multi-positive contrastive loss.

\subsection{Clinical Report Generation}

To enable alignment between retinal images and systemic risk factors within a vision–language framework, structured questionnaire data were converted into synthetic clinical narratives compatible with pretrained text encoders. Using the LLaMA-3.1 API as the text generation engine, each participant’s tabular risk-factor profile was mapped into a standardized clinical-style summary \cite{Grattafiori2024Llama3}. For each subject, the LLM received a predefined documentation template, the subject’s structured demographic, behavioral, cognitive, and lifestyle variables, and explicit instructions to generate a concise report without inferring missing values. The template was adapted from the ``Patient Information'' section of the CARE clinical case reporting guidelines, ensuring consistency with established medical documentation conventions \cite{Gagnier2013CARE}. To minimize variability and preserve numerical fidelity, the prompt enforced a one-to-one mapping between tabular entries and template fields, with unavailable values explicitly marked rather than imputed. This controlled translation process produces semantically enriched text representations that enable structured health information to be embedded within a shared multi-modal latent space.

\subsection{Image and Text Encoders}

Let $\mathbf{x}_p$ denote the retinal image and $\mathbf{t}_p$ the associated clinical report for subject $p$.
The image encoder $E_I(\cdot)$ and text encoder $E_T(\cdot)$ produce modality-specific embeddings.
In our implementation, we instantiate $E_I$ as RETFound \cite{Zhou2023RETFound} and $E_T$ as GatorTron \cite{Yang2022GatorTron}. 
Each encoder is followed by a lightweight linear projection layer to map features into a shared embedding space of dimension $d$.

\begin{equation}
\mathbf{z}^I_p = E_I(\mathbf{x}_p), 
\qquad
\mathbf{z}^T_p = E_T(\mathbf{t}_p).
\end{equation}

Both embeddings are projected into a shared $d$-dimensional space and $\ell_2$-normalized, yielding
$\hat{\mathbf{z}}^I_p = \mathbf{z}^I_p / \lVert \mathbf{z}^I_p \rVert_2$
and
$\hat{\mathbf{z}}^T_p = \mathbf{z}^T_p / \lVert \mathbf{z}^T_p \rVert_2$.
A learnable logit scale parameter $s$ controls the contrastive temperature,
with $\tau = \exp(-s)$.

\subsection{Intra-Modality Similarity for Phenotypic Grouping}

To capture phenotypic similarity between subjects, we construct intra-modality similarity matrices based on cosine similarity between normalized representations. Let $\hat{\mathbf{z}}^I_p$ denote the normalized image embedding produced by the image encoder, and let $\hat{\mathbf{z}}^T_p$ denote the normalized text-derived risk-profile embedding for subjects $p,q$.

\begin{equation}
S_{ii}(p,q) = \langle \hat{\mathbf{z}}^I_p, \hat{\mathbf{z}}^I_q \rangle
\end{equation}
\begin{equation}
S_{tt}(p,q) = \langle \hat{\mathbf{z}}^T_p, \hat{\mathbf{z}}^T_q \rangle
\end{equation}

Here, $S_{ii}$ captures similarity in the learned retinal image embeddings, while $S_{tt}$ captures similarity between clinical report embeddings.

\subsection{Differentiable Phenotypic Weighting}
We transform these similarities into soft membership signals using sigmoid gating with thresholds $\tau_F,\tau_T$ and learnable sharpness parameters $g_F,g_T$:
\begin{equation}
a_F(p,q) = \sigma\!\left(\frac{S_{ii}(p,q)-\tau_F}{g_F}\right),
\qquad
a_T(p,q) = \sigma\!\left(\frac{S_{tt}(p,q)-\tau_T}{g_T}\right),
\end{equation}
where $\sigma(\cdot)$ denotes the logistic sigmoid function. 
Finally, we combine the two signals using a differentiable probabilistic union operator to obtain the phenotypic weighting score
\begin{equation}
W_{pq} = 1 - \bigl(1-a_F(p,q)\bigr)\bigl(1-a_T(p,q)\bigr),
\qquad
W_{pq}\in[0,1].
\end{equation}
Pairs with larger $W_{pq}$ are treated as more strongly aligned in phenotype space and receive higher weight as positives in the multi-positive contrastive objective.

\subsection{Phenotypic Similarity-Weighted Multi-Positive Contrastive Loss}

Cross-modal similarity between image and text embeddings is defined as:
\begin{equation}
S_{it}(p,q) = \left\langle \hat{\mathbf{z}}^I_p, \hat{\mathbf{z}}^T_q \right\rangle.
\end{equation}
Logits are computed using temperature scaling with a learnable log-temperature parameter $s$ and a learnable bias term $\beta$::
\begin{equation}
\ell_{pq} = \frac{S_{it}(p,q)}{\tau} - \beta,
\qquad
\tau = \exp(-s).
\end{equation}
We optimize a soft-target multi-positive contrastive objective:
\begin{equation}
\mathcal{L}_{\mathrm{MP}} 
= \frac{1}{N^2} 
\sum_{p=1}^{N} \sum_{q=1}^{N}
\left[
W_{pq}\,\log\!\bigl(1+\exp(-\ell_{pq})\bigr)
+
\bigl(1-W_{pq}\bigr)\,\log\!\bigl(1+\exp(\ell_{pq})\bigr)
\right].
\end{equation}
When $W_{pq}$ approaches 1, the pair $(p,q)$ is treated as a positive match, when $W_{pq}$ approaches 0, it is treated as a negative pair. 
Intermediate values allow soft supervision based on phenotypic similarity.

\section{Experiments}

\subsection{Dataset and Preprocessing}

\begin{table}[!t]
\centering
\caption{Cohort characteristics across data splits.}
\begin{tabular}{lccc}
\toprule
 & \textbf{Train} & \textbf{Validation} & \textbf{Test} \\
 & (n=30,462) & (n=3,384) & (n=5,396) \\
\midrule
Gender: (male \%) & 45.10 & 45.41 & 45.10 \\
Age: mean (s.d) & 55.53 (8.24) & 55.78 (8.12) & 55.52 (8.17) \\
Ethnicity: (British \%) & 84.08 & 83.51 & 88.51 \\
\bottomrule
\end{tabular}
\end{table}
A comprehensive set of demographic, behavioral, cognitive, and lifestyle variables was extracted from the UK Biobank \cite{Bycroft2018UKBiobank} baseline assessment and compiled as candidate risk factors based on established epidemiological and biomarker evidence linking modifiable exposures to Alzheimer’s disease and dementia risk \cite{Leshner2017Prevention,Suzgun2025SleepADRisk,Xiong2023RiskReview,Hayden2024ModifiableRisk,Huszar2024Progression,Livingston2024Lancet}. These include factors associated with amyloid and tau pathology, sleep disturbance, cardiometabolic health, and other modifiable determinants of neurodegeneration. 

Color fundus photographs (CFPs) from the initial UK Biobank assessment visit were used for image-based modeling. Images underwent automated quality control to exclude low-quality scans, retaining only high-quality CFPs for downstream analysis \cite{Zhou2022AutoMorph}. Preprocessed CFPs are input into a RETFound-initialized vision encoder, which is fine-tuned during training \cite{Zhou2023RETFound}. Each image was resized to match the input resolution of the pretrained RETFound encoder and normalized using standard channel-wise mean and standard deviation values consistent with its pretraining setup. To ensure consistent anatomical orientation across subjects, right-eye images were horizontally flipped prior to encoding.

\subsection{Implementation Details}

RETFound and GatorTron were used as image and text encoders. The vision encoder is initialized with RETFound weights and fine-tuned end-to-end, while the text encoder is kept frozen. Lightweight linear projections map both modalities into a shared $d=1024$-dimensional space. Embeddings were $\ell_2$-normalized prior to similarity computation. The batch size was 128.
Optimization used AdamW with hyperparameters selected via Optuna \cite{Akiba2019Optuna}. The final learning rate was $2.42\times10^{-4}$, $\epsilon=8.61\times10^{-7}$, and weight decay $0.0232$. 
Phenotypic similarity thresholds were initialized from empirical intra-modality cosine similarity distributions computed on 85\% of the development set, restricting the search to the upper interquartile range. 

\section{Results}
To evaluate the proposed framework, we compared our method against strong retinal and biomedical foundation models. We included RETFound \cite{Zhou2023RETFound}, a large-scale retinal image foundation model; RET-CLIP \cite{Du2024RETCLIP}, a retinal image–text contrastive pretraining framework; and MM-Retinal \cite{Wu2024MMRetinal}, a knowledge-enhanced retinal vision–language model. We additionally evaluated two general biomedical multi-modal foundation models: PMC-CLIP \cite{Lin2023PMCClip} and BiomedCLIP \cite{Zhang2025BiomedCLIP}. 
Because RETFound is an image-only encoder, we paired it with GatorTron \cite{Yang2022GatorTron} to construct multi-modal representations, concatenating image and text embeddings for downstream classification. In addition to vision–language baselines, we trained tabular SVM models using structured clinical variables and CFP-derived image features
to assess whether performance gains stem from semantic narrative modeling or solely from image foundation representations. 
All methods were evaluated under an identical multi-modal SVM protocol. Each experiment was repeated across 10 random seeds, and we report mean $\pm$ standard deviation performance.

\begin{table}[t]
\centering
\caption{Comparison of multi-modal and baseline methods for incident AD prediction. Results are reported as mean $\pm$ standard deviation across folds. \textbf{Bold} and \underline{Underline} represent the best and the second best results.}
\label{tab:main_results}
\begin{tabularx}{\linewidth}{>{\raggedright\arraybackslash}Xcccc}
\toprule
 & AUROC & Balanced Accuracy & F1-Score & MCC \\
\midrule
Baseline SVM & 0.593$\pm$0.068 & 0.574$\pm$0.083 & 0.140$\pm$0.089 & 0.076$\pm$0.099 \\

KeepFIT-CFP & 0.490$\pm$0.063 & 0.505$\pm$0.0412 & 0.099$\pm$0.034 & 0.002$\pm$0.046 \\

BiomedCLIP & 0.525$\pm$0.064 & 0.522$\pm$0.060 & 0.121$\pm$0.052 & 0.023$\pm$0.054 \\

RETCLIP & 0.558$\pm$0.076 & 0.527$\pm$0.042 & 0.106$\pm$0.069 & 0.028$\pm$0.051 \\

PMC-CLIP & 0.471$\pm$0.049 & 0.484$\pm$0.020 & 0.076$\pm$0.023 & -0.022$\pm$0.023 \\

RETFound + GatorTron & 0.642$\pm$0.052 & 0.581$\pm$0.069 & 0.185$\pm$0.099 & 0.119$\pm$0.101 \\

REVEAL (no GACL) & 0.654$\pm$0.092 & 0.602$\pm$0.075 & 0.205$\pm$0.096 & 0.144$\pm$0.105 \\

REVEAL (with GACL) & \underline{0.658$\pm$0.090} & \underline{0.609$\pm$0.079} & \underline{0.207$\pm$0.100} & \underline{0.146$\pm$0.111} \\
\textbf{REVEAL++} &
\textbf{0.678$\pm$0.061} &
\textbf{0.613$\pm$0.048} &
\textbf{0.236$\pm$0.079} &
\textbf{0.168$\pm$0.088} \\
\bottomrule
\end{tabularx}
\end{table}

In the incident AD prediction task (Table~\ref{tab:main_results}), our phenotypic-weighted multi-positive contrastive framework consistently outperformed all comparison methods, indicating that soft, differentiable phenotypic alignment leads to more coherent multi-modal representations. Rather than relying on single positive pairs or hard grouping, the proposed formulation allows subjects with similar risk profiles to contribute proportionally during training, yielding stronger downstream discrimination. While pretrained vision–language baselines such as RETFound+GatorTron and RET-CLIP capture meaningful retinal–text correspondences, they do not explicitly model phenotypic structure, which may be important for long-horizon neurodegenerative risk prediction.

These findings suggest that modeling phenotypic similarity as a continuous, differentiable weighting mechanism enables smoother transitions between positive and negative supervision, leading to more coherent multi-modal embedding spaces and improved long-horizon neurodegenerative risk prediction.

\section{Discussion}
Alzheimer’s disease is  increasingly understood not as a binary condition but a long-term neurodegenerative process that evolves over years prior to diagnosis. Pathological changes including amyloid deposition, tau accumulation, vascular dysfunction, and systemic metabolic dysregulation emerge progressively and interact across multiple biological scales before cognitive symptoms become apparent \cite{Chow2026NeurodevelopmentalOrigins,Livingston2024Lancet}. Retinal microvascular alterations and structural remodeling likewise develop along a continuum, reflecting cumulative exposure to systemic and neurodegenerative risk factors. Therefore, similarity between individuals along disease-relevant dimensions is continuous rather than discretely separable.

Hard similarity thresholds impose artificial boundaries on this biological continuum by assigning subjects to fixed phenotypic groups. While such grouping can strengthen contrastive supervision, it implicitly assumes well-defined subtype partitions that may not reflect the underlying progression of the disease. However, such discrete subtype boundaries may not exist during the preclinical stages of neurodegenerative disease, where pathological processes evolve gradually and heterogeneously across individuals. This mismatch can limit the ability of representation learning methods to capture subtle transitions in risk states. 
In contrast, the proposed differentiable phenotypic weighting mechanism allows phenotypic similarity to modulate supervision strength continuously. Participants with partially overlapping risk profiles or subtly similar retinal signatures contribute proportionally during training, enabling smoother organization of the shared embedding space while preserving meaningful inter-subject variation.

This formulation more closely reflects the pathophysiology of preclinical AD, where risk accumulates gradually and manifests heterogeneously across individuals \cite{Jack2018NIAAA}. By relaxing discrete grouping into continuous supervision, the model is able to represent intermediate phenotypic states that may correspond to early pathological changes. The resulting embedding geometry reflects a continuum of risk rather than rigid clusters, providing a representation that is both biologically plausible and better suited for early risk stratification.


\section{Conclusion}
We presented REVEAL\Plus, a differentiable phenotypic alignment framework for multi-modal learning from retinal imaging and clinical risk narratives in preclinical Alzheimer’s disease prediction. By replacing discrete threshold-based grouping with continuous similarity-driven supervision, the proposed approach enables phenotypic relationships to be learned jointly with representation alignment, allowing population structure to emerge directly from data. This formulation better captures the gradual and heterogeneous nature of neurodegenerative disease progression and leads to improved risk prediction performance. More broadly, differentiable phenotypic alignment offers a strategy for modeling structured variability in multi-modal biomedical data, with potential applications spanning chronic disease risk prediction, precision medicine, longitudinal health modeling, and large-scale population health analytics across diverse clinical domains.
%
%
%
\bibliographystyle{splncs04}
\bibliography{reveal}

@article{Chow2026NeurodevelopmentalOrigins,
  author  = {Chow, Kim Hei-Man and Abel, Ted},
  title   = {Neurodevelopmental origins of age-related neurodegenerative diseases},
  journal = {eBioMedicine},
  volume  = {124},
  pages   = {106151},
  year    = {2026},
  doi     = {10.1016/j.ebiom.2026.106151},
}

@article{Banna2024ImagingEyeBrain,
  author  = {Banna, HU and Slayo, M and Armitage, JA and Del Rosal, B and Vocale, L and Spencer, SJ},
  title   = {Imaging the eye as a window to brain health: frontier approaches and future directions},
  journal = {Journal of Neuroinflammation},
  year    = {2024},
  volume  = {21},
  number  = {1},
  pages   = {309},
  doi     = {10.1186/s12974-024-03304-3},
  pmid    = {39614308},
  pmcid   = {PMC11606158}
}

@article{BuenoLopez2025CardiometabolicDementia,
  author  = {Bueno Lopez, C and Iona, A and Avery, D and Turnbull, I and Yang, L and Du, H and Chen, Y and Zhang, N and Chen, J and Pei, P and Lv, J and Yu, C and Sun, D and Li, L and Bennett, D and van Duijn, C and Clarke, R and Chen, Z and Bragg, F},
  title   = {Cardiometabolic health and risk of dementia and brain atrophy: a community-based prospective cohort study of 0.5 million adults in China},
  journal = {The Lancet Regional Health -- Western Pacific},
  year    = {2025},
  volume  = {64},
  pages   = {101743},
  doi     = {10.1016/j.lanwpc.2025.101743},
  pmid    = {41341656},
  pmcid   = {PMC12670887}
}

@article{Suzgun2025SleepADRisk,
  author  = {Aktan Süzgün, M and Tang, Q and Stefani, A},
  title   = {Sleep Abnormalities and Risk of Alzheimer's Disease},
  journal = {Current Neurology and Neuroscience Reports},
  year    = {2025},
  volume  = {25},
  number  = {1},
  pages   = {67},
  doi     = {10.1007/s11910-025-01451-5},
  pmid    = {41082137},
  pmcid   = {PMC12518432}
}

@inproceedings{Leem2026REVEAL,
  title     = {{REVEAL}: Multimodal Vision--Language Alignment of Retinal Morphometry and Clinical Risks for Incident {AD} and Dementia Prediction},
  author    = {Leem, Seowung and Gu, Lin and You, Chenyu and Gong, Kuang and Fang, Ruogu},
  booktitle = {Medical Imaging with Deep Learning},
  year      = {2026},
  note      = {Accepted by MIDL 2026. Proceedings of Machine Learning Research (PMLR)},
  url       = {https://openreview.net/pdf?id=aOKAXRHXVw}
}

@article{Bycroft2018UKBiobank,
  author  = {Bycroft, Clare and Freeman, Colin and Petkova, Desislava and Band, Gavin and Elliott, Lloyd T. and Sharp, Kevin and Motyer, Allan and Vukcevic, Damjan and Delaneau, Olivier and O'Connell, Jared and Cortes, Adrian and Welsh, Samantha and Young, Alan and Effingham, Mark and McVean, Gilean and Leslie, Stephen and Allen, Naomi and Donnelly, Peter and Marchini, Jonathan},
  title   = {The UK Biobank resource with deep phenotyping and genomic data},
  journal = {Nature},
  year    = {2018},
  volume  = {562},
  number  = {7726},
  pages   = {203--209},
  doi     = {10.1038/s41586-018-0579-z}
}

@article{Grattafiori2024Llama3,
  title   = {The Llama 3 Herd of Models},
  author  = {Grattafiori, Aaron and Dubey, Abhimanyu and Jauhri, Abhinav and others},
  year    = {2024},
  journal = {arXiv preprint arXiv:2407.21783},
  url     = {https://arxiv.org/abs/2407.21783}
}

@article{Gagnier2013CARE,
  title   = {The CARE Guidelines: Consensus-Based Clinical Case Reporting Guideline Development},
  author  = {Gagnier, Joel J. and Kienle, Gunver and Altman, Douglas G. and Moher, David and Sox, Harold and Riley, David and CARE Group},
  journal = {Global Advances in Health and Medicine},
  volume  = {2},
  number  = {5},
  pages   = {38--43},
  year    = {2013},
  month   = {September},
  doi     = {10.7453/gahmj.2013.008},
  publisher = {SAGE Publications}
}

@article{Zhou2023RETFound,
  title     = {A foundation model for generalizable disease detection from retinal images},
  author    = {Zhou, Yukun and Chia, Mark A. and Wagner, Siegfried K. and Ayhan, Murat S. and Williamson, Dominic J. and Struyven, Robbert R. and Liu, Timing and Xu, Moucheng and Lozano, Mateo G. and Woodward-Court, Peter and Kihara, Yuka and Altmann, Andre and Lee, Aaron Y. and Topol, Eric J. and Denniston, Alastair K. and Alexander, Daniel C. and Keane, Pearse A.},
  journal   = {Nature},
  volume    = {622},
  number    = {7981},
  pages     = {156--163},
  year      = {2023},
  month     = {October},
  doi       = {10.1038/s41586-023-06555-x},
  publisher = {Nature Publishing Group}
}

@article{Yang2022GatorTron,
  title     = {A large language model for electronic health records},
  author    = {Yang, Xi and Chen, Aokun and PourNejatian, Nima and Shin, Hoo Chang and Smith, Kaleb E. and Parisien, Christopher and Compas, Colin and Martin, Cheryl and Costa, Anthony B. and Flores, Mona G. and Zhang, Ying and Magoc, Tanja and Harle, Christopher A. and Lipori, Gloria and Mitchell, Duane A. and Hogan, William R. and Shenkman, Elizabeth A. and Bian, Jiang and Wu, Yonghui},
  journal   = {npj Digital Medicine},
  volume    = {5},
  number    = {1},
  pages     = {1--9},
  year      = {2022},
  month     = {December},
  doi       = {10.1038/s41746-022-00742-2},
  publisher = {Nature Publishing Group}
}

@article{Xiong2023RiskReview,
  title   = {Review of Risk Factors Associated With Biomarkers for Alzheimer Disease},
  author  = {Xiong, Jiayue and Bhimani, Rozina and Carney-Anderson, Lisa},
  journal = {Journal of Neuroscience Nursing},
  volume  = {55},
  number  = {3},
  pages   = {103--109},
  year    = {2023},
  month   = {June},
  doi     = {10.1097/JNN.0000000000000705}
}

@book{Leshner2017Prevention,
  title     = {Preventing Cognitive Decline and Dementia: A Way Forward},
  editor    = {Leshner, Alan I. and Landis, Story and Stroud, Clare and Downey, Autumn},
  publisher = {National Academies Press},
  address   = {Washington, DC},
  year      = {2017},
  month     = {September},
  doi       = {10.17226/24782}
}

@article{Hayden2024ModifiableRisk,
  title   = {Association between Modifiable Risk Factors and Levels of Blood-Based Biomarkers of Alzheimer’s and Related Dementias in the Look AHEAD Cohort},
  author  = {Hayden, Kathleen M. and Mielke, Michelle M. and Evans, Jennifer K. and Neiberg, Rebecca and Molina-Henry, D. and Culkin, M. and Marcovina, S. and Johnson, Karen C. and Carmichael, Owen T. and Rapp, Stephen R. and Sachs, B. C. and Ding, J. and Shappell, H. and Wagenknecht, Lynne and Luchsinger, Jose A. and Espeland, Mark A.},
  journal = {JAR Life},
  volume  = {13},
  pages   = {1--21},
  year    = {2024},
  month   = {January},
  doi     = {10.14283/jarlife.2024.1}
}

@article{Huszar2024Progression,
  title   = {Association of modifiable risk factors with progression to dementia in relation to amyloid and tau pathology},
  author  = {Husz{\'a}r, Zsolt and Solomon, Alina and Engh, Marie Anne and Koszov{\'a}cz, Vanda and Terebessy, Tam{\'a}s and Moln{\'a}r, Zsolt and Hegyi, P{\'e}ter and Horv{\'a}th, Andr{\'a}s and Mangialasche, Francesca and Kivipelto, Miia and Csukly, G{\'a}bor},
  journal = {Alzheimer's Research \& Therapy},
  volume  = {16},
  pages   = {238},
  year    = {2024},
  month   = {October},
  doi     = {10.1186/s13195-024-01602-9}
}

@article{Livingston2024Lancet,
  title   = {Dementia prevention, intervention, and care: 2024 report of the Lancet Standing Commission},
  author  = {Livingston, Gill and Huntley, Jonathan and Liu, Kathy Y. and Costafreda, Sergi G. and Selb{\ae}k, Geir and Alladi, Suvarna and Ames, David and Banerjee, Sube and Burns, Alistair and Brayne, Carol and Fox, Nick C. and Ferri, Cleusa P. and Gitlin, Laura N. and Howard, Robert and Kales, Helen C. and Kivim{\"a}ki, Mika and Larson, Eric B. and Nakasujja, Noeline and Rockwood, Kenneth and Samus, Quincy and Shirai, Kokoro and Singh-Manoux, Archana and Schneider, Lon S. and Walsh, Sebastian and Yao, Yao and Sommerlad, Andrew and Mukadam, Naaheed},
  journal = {The Lancet},
  volume  = {404},
  number  = {10452},
  pages   = {572--628},
  year    = {2024},
  month   = {August},
  doi     = {10.1016/S0140-6736(24)01296-0}
}

@article{Du2024RETCLIP,
  title   = {RET-CLIP: A Retinal Image Foundation Model Pre-trained with Clinical Diagnostic Reports},
  author  = {Du, Jiawei and Guo, Jia and Zhang, Weihang and Yang, Shengzhu and Liu, Hanruo and Li, Huiqi and Wang, Ningli},
  journal = {arXiv preprint arXiv:2405.14137},
  year    = {2024}
}

@article{Wu2024MMRetinal,
  title   = {MM-Retinal: Knowledge-Enhanced Foundational Pretraining with Fundus Image-Text Expertise},
  author  = {Wu, Ruiqi and Zhang, Chenran and Zhang, Jianle and Zhou, Yi and Zhou, Tao and Fu, Huazhu},
  journal = {arXiv preprint arXiv:2405.11793},
  year    = {2024}
}

@article{Zhang2025BiomedCLIP,
  title   = {BiomedCLIP: A Multimodal Biomedical Foundation Model Pretrained from Scientific Image-Text Pairs},
  author  = {Zhang, Sheng and Xu, Yanbo and Usuyama, Naoto and Xu, Hanwen and Bagga, Jaspreet and Tinn, Robert and Preston, Sam and Rao, Rajesh and Wei, Mu and Valluri, Naveen and Wong, Cliff and Tupini, Andrea and Wang, Yu and Mazzola, Matt and Shukla, Swadheen and Liden, Lars and Gao, Jianfeng and Crabtree, Angela and Piening, Brian and Bifulco, Carlo and Lungren, Matthew P. and Naumann, Tristan and Wang, Sheng and Poon, Hoifung},
  journal = {arXiv preprint arXiv:2303.00915},
  year    = {2025}
}

@article{Lin2023PMCClip,
  title   = {PMC-CLIP: Contrastive Language-Image Pre-training using Biomedical Documents},
  author  = {Lin, Weixiong and Zhao, Ziheng and Zhang, Xiaoman and Wu, Chaoyi and Zhang, Ya and Wang, Yanfeng and Xie, Weidi},
  journal = {arXiv preprint arXiv:2303.07240},
  year    = {2023}
}

@inproceedings{Akiba2019Optuna,
  author    = {Akiba, Takuya and Sano, Shotaro and Yanase, Toshihiko and Ohta, Takeru and Koyama, Masanori},
  title     = {Optuna: A Next-generation Hyperparameter Optimization Framework},
  booktitle = {Proceedings of the 25th ACM SIGKDD International Conference on Knowledge Discovery \& Data Mining (KDD '19)},
  pages     = {2623--2631},
  year      = {2019},
  publisher = {ACM},
  address   = {New York, NY, USA},
  doi       = {10.1145/3292500.3330701}
}

@article{Jack2018NIAAA,
  author = {Jack, Clifford R. Jr. and Bennett, David A. and Blennow, Kaj and Carrillo, Maria C. and Dunn, Bruce and Haeberlein, Stephen B. and Holtzman, David M. and Jagust, William and Jessen, Frank and Karlawish, Jason and Liu, Eric and Molinuevo, Jos{\'e} L. and Montine, Thomas and Phelps, Creighton and Rankin, Katherine P. and Rowe, Christopher C. and Scheltens, Philip and Siemers, Eric and Snyder, Heather M. and Sperling, Reisa},
  title = {NIA-AA Research Framework: Toward a biological definition of Alzheimer's disease},
  journal = {Alzheimer's \& Dementia},
  volume = {14},
  number = {4},
  pages = {535--562},
  year = {2018},
  doi = {10.1016/j.jalz.2018.02.018}
}

@article{Zhou2022AutoMorph,
  title        = {AutoMorph: Automated Retinal Vascular Morphology Quantification Via a Deep Learning Pipeline},
  author       = {Zhou, Yukun and Wagner, Siegfried K. and Chia, Mark A. and Zhao, An and Woodward-Court, Peter and Xu, Moucheng and Struyven, Robbert and Alexander, Daniel C. and Keane, Pearse A.},
  journal      = {Translational Vision Science \& Technology},
  volume       = {11},
  number       = {7},
  pages        = {12},
  year         = {2022},
  month        = {July},
  doi          = {10.1167/tvst.11.7.12},
  url          = {https://doi.org/10.1167/tvst.11.7.12},
  issn         = {2164-2591}
}

\end{document}